\documentclass[10pt,twocolumn,letterpaper]{article}

\usepackage{cvpr}              

\usepackage{amsmath,amssymb,amsfonts}
\usepackage{algorithmic}
\usepackage{graphicx}
\usepackage{textcomp}
\usepackage[dvipsnames]{xcolor}

\usepackage{algorithmic}
\usepackage{amsmath}
\usepackage{graphicx}
\usepackage{color} 
\usepackage{algorithm} 
\usepackage[font=scriptsize]{subcaption}
\usepackage{multirow}
\usepackage{multicol}
\usepackage{tabularx,booktabs}
\usepackage{url}

\newcommand{\boldhdr}[1]{\smallskip \noindent \textbf{#1.}}

\usepackage{comment}
\usepackage{ifthen}
\newcommand{\showcomments}{yes}
\newcommand\kai[1]{
    \ifthenelse{\equal{\showcomments}{yes}}{{\color{blue} [Note: #1]}}{\ignorespaces}
}

\definecolor{cvprblue}{rgb}{0.21,0.49,0.74}
\usepackage[pagebackref,breaklinks,colorlinks,allcolors=cvprblue]{hyperref}

\def\paperID{*****} 
\def\confName{CVPR}
\def\confYear{2025}

\title{Progressive Element-wise Gradient Estimation for Neural Network Quantization}

\author{Kaiqi Zhao\\
Oakland University\\
{\tt\small kaiqizhao@oakland.edu}
}

\begin{document}
\maketitle

\begin{abstract}
Neural network quantization aims to reduce the bit-widths of weights and activations, making it a critical technique for deploying deep neural networks on resource-constrained hardware. Most Quantization-Aware Training (QAT) methods rely on the Straight-Through Estimator (STE) to address the non-differentiability of discretization functions by replacing their derivatives with that of the identity function. While effective, STE overlooks discretization errors between continuous and quantized values, which can lead to accuracy degradation --- especially at extremely low bit-widths. 
In this paper, we propose Progressive Element-wise Gradient Estimation (PEGE), a simple yet effective alternative to STE, which can be seamlessly integrated with any forward propagation methods and improves the quantized model accuracy.
PEGE progressively replaces full-precision weights and activations with their quantized counterparts via a novel logarithmic curriculum-driven mixed-precision replacement strategy.
Then it formulates QAT as a co-optimization problem that simultaneously minimizes the task loss for prediction and the discretization error for quantization, providing a unified and generalizable framework. 
Extensive experiments on CIFAR-10 and ImageNet across various architectures (e.g., ResNet, VGG) demonstrate that PEGE consistently outperforms existing backpropagation methods and enables low-precision models to match or even outperform the accuracy of their full-precision counterparts.
\end{abstract}

\section{Introduction}
Deep neural networks (DNNs) have substantial computational and memory requirements. As the use of deep learning grows rapidly on a wide variety of Internet of Things and devices, the mismatch between resource-hungry DNNs and resource-constrained devices also becomes increasingly severe~\citep{zhang2023towards, zhao2023automatic}.
Quantization is one of the important model compression approaches to address this challenge by converting the full-precision model weights or activations to lower precision.
In particular, Quantization-Aware Training (QAT)~\citep{krishnamoorthi2018quantizing} has achieved promising results in creating low-bit models, which starts with a pre-trained model and performs quantization during retraining. 
However, existing QAT methods often suffer from noticeable accuracy degradation, especially when targeting extremely low-bit precision. 
Moreover, no algorithm achieves consistent performance on every model architecture (e.g., VGG, ResNet, MobileNet)~\citep{li2021mqbench}.
Most QAT approaches rely on the Straight-Through Estimator (STE)~\cite{bengio2013estimating} to address the fundamental challenge of non-differentiability in the discretization step. 
Specifically, the round function used in quantization has zero or undefined derivatives, which leads to vanishing or exploding gradients during backpropagation. 
STE circumvents this by approximating the gradient as an identity function, enabling stable training. 
Despite its simplicity and widespread adoption, STE overlooks discretization errors between latent and discrete values, which become significant at low bit-widths, leading to substantial accuracy loss.
To mitigate this issue, Element-Wise Gradient Scaling (EWGS)~\cite{lee2021network} was proposed to minimize the discretization error during backpropagation by scaling gradients according to the difference between continuous and discrete values. 
Although EWGS demonstrates improved results, its effectiveness remains tightly coupled with its specific forward propagation method (see Section~\ref{sec:evaluation_results} for detailed discussions), limiting its flexibility and generalizability.
Therefore, we argue that there is a critical need for a simple, effective, and generalizable gradient estimation method that enhances training stability, mitigates accuracy degradation from reduced precision, and seamlessly integrates with any forward method to consistently improve overall quantization performance.


In this paper, we propose Progressive Element-wise Gradient Estimation (PEGE), a novel backward propagation method for QAT.
PEGE first unifies the forward dynamics of diverse quantizers and progressively replaces full-precision weights with their low-precision counterparts during training.
The replacement process is controlled by a novel logarithmic Curriculum Learning driven strategy which updates replacing rates with a dynamic, smoothing mechanism.
PEGE then minimizes the discretization error to ensure that estimated gradients closely approximate those of the full-precision model. 
PEGE further formulates QAT as a co-optimization problem that simultaneously minimizes the task loss and the discretization error between full-precision weights/activations and their quantized counterparts, offering a unified and theoretically grounded perspective on quantization training.


Our comprehensive evaluation shows that PEGE significantly outperforms the state-of-the-art backpropagation methods on widely used model architectures (VGG and ResNet) across CIFAR-10 and ImageNet datasets.
Notably, PEGE enables low-precision models to match or even surpass the accuracy of their full-precision counterparts.
For example, on ResNet-20 with CIFAR-10, PEGE converges faster and achieves higher accuracy than EWGS and STE by 0.97\% and 0.45\%, respectively.
On ImageNet, PEGE improves the full-precision model's accuracy by 0.21\%, whereas STE reduces accuracy by 0.5\% to 1.8\%, and EWGS results in a 0.27\% accuracy drop.

\section{Background and Related Works}

There are two main approaches to quantization.
Post-Training Quantization (PTQ)~\citep{li2021brecq} quantizes a pre-trained model without retraining and often leads to worse accuracy degradation than QAT~\citep{krishnamoorthi2018quantizing}, which performs quantization during retraining. 
This paper focuses on QAT.
Many QAT works emphasize designing the forward and backward propagation of the quantizer, the function that converts continuous weights or activations to discrete values.
Early works such as BNN~\citep{courbariaux2016binarized} and XNOR-Net~\citep{rastegari2016xnor} employ channel-wise scaling in the forward pass, while DoReFa-Net~\citep{zhou2016dorefa} introduces a universal scalar for all filters.
Recent works utilize trainable parameters for the quantizer, improving control over facets such as clipping ranges (e.g., PACT~\citep{choi2018pact}, LSQ~\citep{esser2019learned}, APoT~\citep{li2019additive}, and DSQ~\citep{gong2019differentiable}), i.e., the bounds where the input values are constrained, and quantization intervals (e.g., QIL~\citep{jung2019learning} and EWGS~\citep{lee2021network}), i.e., the step size between adjacent quantization levels.
However, existing QAT methods lead to different levels of accuracy loss and are motivated by various heuristics, lacking a commonly agreed theory.
Furthermore, MQbench~\citep{li2021mqbench} reveals that the difference between QAT algorithms is not as substantial as reported in their original papers;
and no algorithm achieves the best performance on all architectures. 
%
Therefore, this paper aims to close this gap, providing a simple, effective, and generalizable gradient estimation method that seamlessly integrates with any forward method and consistently improves overall quantization performance across both low-bit and high-bit precisions.


\section{Methodology}


\subsection{Forward Propagation}
Let us define $Quant(\cdot)$ as a uniform quantizer that converts a full-precision input $x$ to a quantized output $x_q = Quant(x)$.
$x$ can be the activations or weights of the network.
First, the quantizer $Quant(\cdot)$ applies a clipping function $Clip(\cdot)$, which normalizes and restricts the full-precision input $x$ to a limited range, producing a full-precision latent presentation $x_c$, as follows:
\vspace{-3pt}
\begin{equation}\label{equ:clip_function}
\small
    x_c = Clip(x, \{p_i\}_{i=1}^{i=K_c}, v, m),
\vspace{-3pt}
\end{equation}
where $v$ and $m$ are the lower and upper bounds of the range, respectively,
$\{p_i\}_{i=1}^{i=K_c}$ denotes the set of trainable parameters needed for quantization,
and $K_c$ denotes the number of parameters.
%
Note that different quantizers have different schemes for $Clip(\cdot)$.
For example, in PACT, the lower bound $v$ is set to 0 and the upper bound $m$ is a trainable parameter optimized during training. 
The quantizer requires only one parameter, that is, $\{p_1|p_1=m, K_c=1\}$,
and the clipping function is described as: $x_c = Clip(x, \{p_1|p_1=m\}, 0, m) = 0.5(|x|-|x-m|+m)$.
%
In EWGS, $v$ and $m$ are set to 0 and 1, respectively, and every quantized layer uses separate parameters (i.e. $p_1$ and $p_2$) for quantization intervals: $x_c = Clip(x, \{p_1, p_2\}, 0, 1) = clip(\frac{x-p_1}{p_1-p_2}, 0, 1)$.

Then, the quantizer $Quant(\cdot)$ converts the clipped value $x_c$ to a discrete quantization point $x_q$ using the function $R(\cdot)$ that contains a round function: 
\vspace{-3pt}
\begin{equation}\label{equ:round_function}
\small
    x_q = R(x_c, b, \{q_i\}_{i=1}^{i=K_r}),
\vspace{-3pt}
\end{equation}
where $b$ is the bit width and $\{q_i\}_{i=1}^{i=K_r}$ denotes the set of trainable parameters.
Note that $\{q_i\}_{i=1}^{i=K_r}$ is not necessary for some quantizers. 
For example, in EWGS, if activations are the input, $x_q = R(x_c, b) = \frac{round((2^b-1) \cdot x_c)}{2^b-1}$; and if weights are the input, $x_q = R(x_c, b) = 2(\frac{round((2^b-1) \cdot x_c)}{2^b-1} - 0.5)$.
%
In some quantizers, like PACT and LSQ, the trainable parameters in the function $R(\cdot)$ are the same as those in the clipping function $Clip(\cdot)$, that is, $\{q_i|q_i=p_i\}_{i=1}^{i=K_r}$ and $K_r=K_c$.
%

In summary, the quantizer $Quant(\cdot)$ is described as: $x_q = Quant(x, \alpha, b, v, m)$,
where $\alpha$ denotes a shorthand for the set of all the parameters in the functions $R(\cdot)$ and $Clip(\cdot)$: $\alpha = \{ \{p_i\}_{i=1}^{i=K_c}, \{q_i\}_{i=1}^{i=K_r} \}$.
\subsection{Backward Propagation}
Directly training a quantized network using back-propagation is impossible since the quantizer $Q(\cdot)$ is non-differentiable.
This issue arises due to the round function in Eq.~\ref{equ:round_function}, which produces near-zero derivatives almost everywhere.
To solve this problem, most QAT works use Straight-Through Estimator (STE)~\citep{bengio2013estimating} to approximate the gradients: $\frac{\partial L}{\partial x_c} = \frac{\partial L}{\partial x_q}$.
Instead of the commonly used STE for backpropagation, PEGE uses two core strategies, including a logarithmic curriculum-driven progressive precision replacement, and an adaptive gradient correction based on discretization error.

\boldhdr{Progressive Precision Replacement}
PEGE stochastically mixes different full-precision and quantized weights/activations in each backpropagation step. 
Consider backpropagation step $\mathcal{T}$, we generate an independent Bernoulli random variables, denoted as $r_{\mathcal{T}}$, which is either equal to 1 or 0:
\vspace{-5pt}
\begin{equation}\label{equ:replacing_rate}
        r_{\mathcal{T}} \sim Bernoulli(p)
\vspace{-5pt}
\end{equation}
where $0 < p \leqslant 1$, and $r_{\mathcal{T}} \in \left \{ 0, 1 \right \}$. The replacing rate $p$ controls the probability of $r_{\mathcal{T}}$ being 1. When $r_{\mathcal{T}}$ is equal to 1, the updated parameters come from the quantized parameters, i.e., all the full-precision parameters are replaced.
This mechanism enables quantized weights/activations to progressively receive guidance from their full-precision counterparts through gradient-level interaction.

\begin{figure}[t]
    \centering
	\includegraphics[width=0.7\linewidth]{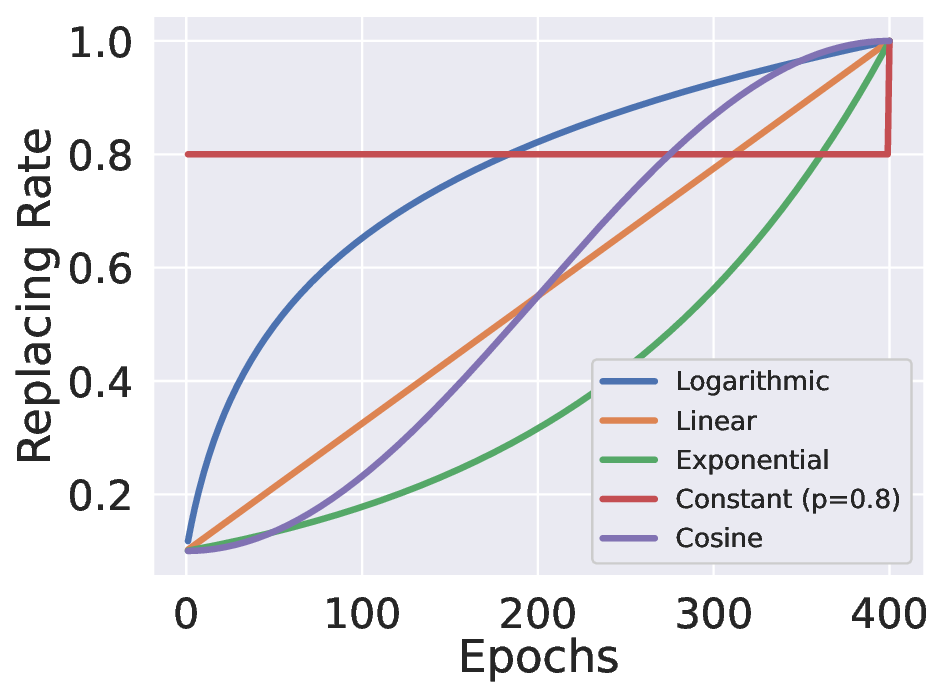}
	\vspace{-10pt}
	\caption{The replacing curves of five replacement schedulers.}
	\label{fig:replace_curves_p}
\end{figure}

We propose a novel logarithmic curriculum replacement strategy to update the replacement rate.
Let $\mathcal{T}$ denote the $\mathcal{T}$-th iteration/training step, so that the replacing rate $p_{\mathcal{T}}$ at iteration $\mathcal{T}$ can be calculated as:
\vspace{-5pt}
\begin{equation}
     p_{\mathcal{T}} = min(log_{B}^{k\mathcal{T}+b}, 1.0)
     \vspace{-5pt}
\end{equation}
where $B$ is the base value of the logarithmic function, $log_{B}^b$ is the basic replacing rate, and $k$ is the coefficient. Figure~\ref{fig:replace_curves_p} illustrates the replacing curves of Constant, Linear, Logarithmic, and Exponential Replacement Schedulers. 

\boldhdr{Adaptive Gradient Estimation}
After progressive replacement, PEGE introduces a novel formula to integrate the discretization error ($x_c - x_q$), representing the deviation between full-precision and its quantized weights/activations:
\begin{equation}\label{equ:back_propogation}
\small
    \frac{\partial L}{\partial x_c} = \frac{\partial L}{\partial x_q} + \mu \cdot (x_c - x_q),
\end{equation}
where $\mu \geq 0 $ is a scaling factor that modulates the influence of the discretization error.
To align this modulation with training dynamics, PEGE uses an Exponential Curriculum Learning driven strategy to update $\mu$: 
$\mu_{\mathcal{T}} = \mu_{\text{max}} \cdot \left(1 - e^{-k \mathcal{\mathcal{T}}}\right)$,
with $\mu_{\text{max}}$ as the maximum value, $k$ controlling the rate of growth, and $\mathcal{T}$ as the iteration step.
\subsection{Optimization Objective}
We frame QAT as an optimization problem minimizing both the discretization error as well as the discrepancy between model predictions and true labels.
The goal is precise quantization without sacrificing predictive accuracy. 
Thus, the optimization objection is defined as follows:
\vspace{-3pt}
\begin{equation}\label{equ:optimization}
    \small
    \begin{split}
    & \quad \quad \quad \underset {W_f, \alpha_W, \alpha_A} {min} L(W_f)\\
    s.t. \quad  &W_q = Quant_W(W_f, \alpha_W, b_W, v_W, m_W) \\
    &A_q = Quant_A(A_f, \alpha_A, b_A, v_A, m_A),  
    \end{split}
\vspace{-6pt}
\end{equation}
where $L(\cdot)$ is the cross entropy loss with labels, $Quant_W(\cdot)$ and $Quant_A(\cdot)$ are the quantizers for weights and activations,
and $W_f$/$A_f$ and $W_q$/$A_q$ are the model's full-precision and quantized weights/activations.

To the best of our knowledge, we are the first to generalize diverse quantizers, encompassing forward and backward propagations into a unified formulation of an optimization problem.
This generalization and formulation enable our backward method to be flexible in integrating various forward methods and effectively improve the overall quantization performance (see Section~\ref{sec:evaluation_results} for the results).

\section{Evaluation}


\boldhdr{Models and Datasets}
We evaluate our proposed method on a variety of image classification models and datasets, including 
1) ResNet-20 and VGG-16 on CIFAR-10 which contains 50K (32$\times$32) RGB training images, belonging to 10 classes, 
and 2) ResNet-18 on ImageNet containing about 1.4M training images and 50K validation images, with 1000 categories.

\boldhdr{Implementation Details} 
We implement all the techniques on PyTorch version 1.10.0 and Python version 3.9.7, and conducted experiments on eight Nvidia V100 GPUs. The learning rate decays to 0.0 using a cosine annealing schedule. Nesterov SGD optimizer is used with a momentum of 0.9 on ImageNet, and Adam optimizer is used for CIFAR-10, respectively. The batch size is set to 64 for CIFAR-10 and to 256 for ImageNet.


\subsection{Results}\label{sec:evaluation_results}

\boldhdr{Results on CIFAR-10}
Table~\ref{tab:results_cifar10} presents the top-1 test accuracy of ResNet-20 and VGG-16 models on CIFAR-10, where activations and weights are quantized to 2 bits using the same forward method, i.e., EWGS, with different backward methods, including STE~\cite{bengio2013estimating}, EWGS~\cite{lee2021network}, and the proposed PEGE.

Our method significantly outperforms 1) EWGS by 0.97\% on ResNet-20 and by 0.11\% on VGG-16, and 2) STE by 0.45\% on ResNet-20 and by 0.21\% on VGG-16, respectively.
In addition, our method enables a 2-bit model to achieve a comparable accuracy (91.62\% vs 91.95\% on ResNet-20 and 93.73\% vs 93.84\% on VGG-16) with its full-precision model with a compression ratio of 16$\times$ (quntized from 32 bits to 2 bits).
\textit{To our best knowledge, the 2-bit ResNet-20 and 2-bit VGG-16 quantized by our method achieve the state-of-the-art top-1 test accuracy.}

\begin{figure}[t]
  \centering
  \begin{subfigure}{.45\linewidth}
    \centering
      \includegraphics[width=3.8cm]{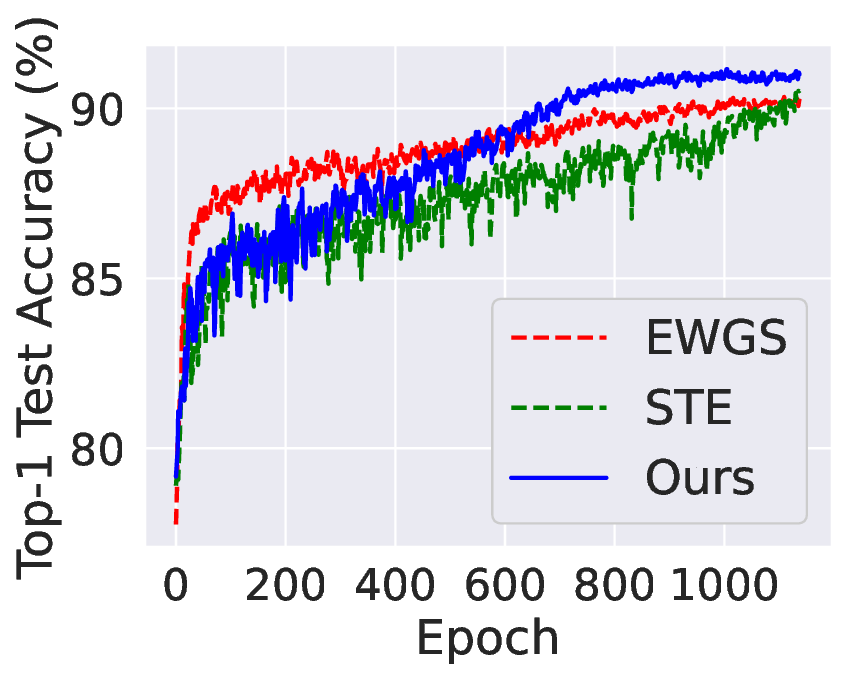}
    \caption{ResNet-20 (W2A2, CIFAR-10)}
    \label{fig:overall_cifar10_resnet20}
  \end{subfigure}
  \begin{subfigure}{.45\linewidth}
    \centering
      \includegraphics[width=3.8cm]{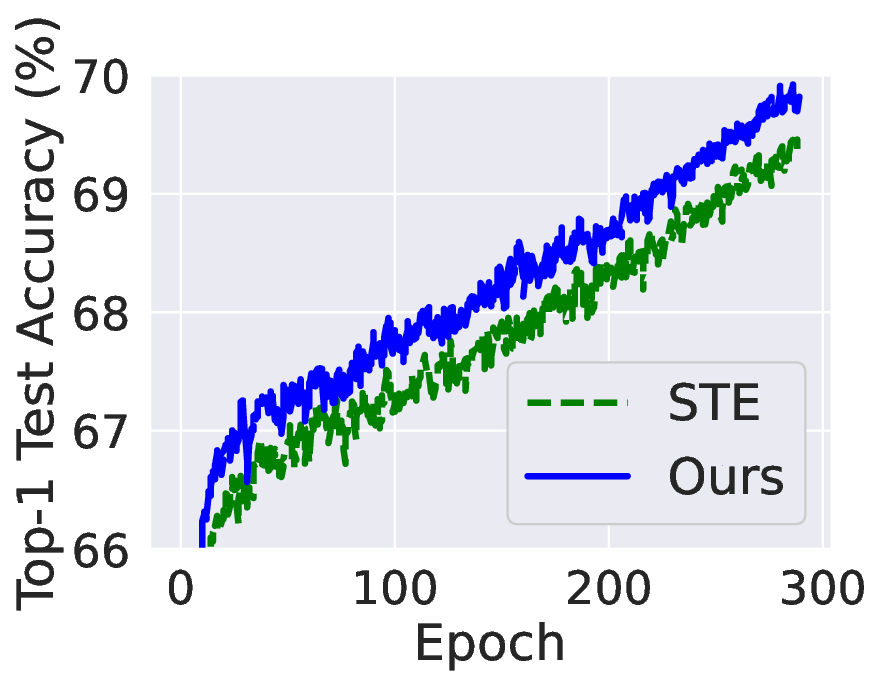}
    \caption{ResNet-18 (W4A4, ImageNet)}
    \label{fig:overall_imagenet_resnet18}
  \end{subfigure}
  \caption{Top-1 test accuracy evolution of a) ResNet-20 (W2A2) on CIFAR-10 and 2) ResNet-18 (W4A4) on ImageNet during training, using different backward methods with the same forward methods (EWGS for CIFAR-10 and PACT for ImageNet).}
   \vspace{-10pt}
\end{figure}

Figure~\ref{fig:overall_cifar10_resnet20} illustrates the top-1 test accuracy evolution for quantized ResNet-20 in each epoch during training on CIFAR-10.
Compared to EWGS and STE, PEGE accelerates the convergence speed of 2-bit ResNet-20. 

\boldhdr{Results on ImageNet}
As shown in Table~\ref{tab:results_imagenet}, our method is the only one that enables a small, 4-bit ResNet-18 to surpass the top-1 test accuracy of its full-precision counterpart on ImageNet. 
Specifically, our method improves the full-precision model's accuracy by 0.21\%, whereas STE reduces accuracy by 0.5\% to 1.8\%, and EWGS results in a 0.27\% accuracy drop.
Moreover, PEGE significantly outperforms EWSG and STE in top-1 test accuracy for 4-bit quantization by 0.48\% and 0.45\%, respectively, using PACT as the forward method. 
Furthermore, although PACT initializes the low-precision model with a much better full-precision model in its original paper (71.0\% vs. 69.8\%), its final accuracy (69.2\%) remains lower than ours (70.01\%).

Figure~\ref{fig:overall_imagenet_resnet18} demonstrates that PEGE enables 4-bit ResNet-20 to converge significantly faster than STE on ImageNet.
Notably, PACT originally employs STE as its backward method.
\textit{These results confirm that simply replacing the commonly used STE with our method can substantially improve both the convergence speed and final accuracy of the whole quantization.}



\begin{table}[t]
\scriptsize
\centering
\caption{Top-1 test accuracy (\%) on CIFAR-10.
``W$*$A$\times$'' denotes that the weights and activations are quantized into $*$bit and $\times$bit, respectively.}
\label{tab:results_cifar10}
\setlength{\tabcolsep}{7.pt}
\vspace{-10pt}
\begin{tabularx}{0.48\textwidth}{@{}c|c|cc|cc@{}}
\toprule
Bit-Width             & Model                      & \multicolumn{1}{c|}{\begin{tabular}[c]{@{}c@{}}Forward \\ Method\end{tabular}} & \begin{tabular}[c]{@{}c@{}}Backward \\ Method\end{tabular} & Accuracy       & \begin{tabular}[c]{@{}c@{}}Accuracy \\ Difference\end{tabular} \\ \midrule
\multirow{8}{*}{W2A2} & \multirow{4}{*}{ResNet-20} & \multicolumn{2}{c|}{Full Precision (FP)}                                                                                                    & 91.95          & -                                                              \\ \cmidrule(l){3-6} 
                      &                            & \multicolumn{1}{c|}{\multirow{3}{*}{EWGS}}                                     & EWGS                                                       & 90.65          & -1.30                                                           \\
                      &                            & \multicolumn{1}{c|}{}                                                          & STE                                                        & 91.17          & -0.78                                                          \\
                      &                            & \multicolumn{1}{c|}{}                                                          & \textbf{Ours}                                                       & \textbf{91.62} & \textbf{-0.33}                                                 \\ \cmidrule(l){2-6} 
                      & \multirow{4}{*}{VGG-16}    & \multicolumn{2}{c|}{Full Precision (FP)}                                                                                                    & 93.84          & -                                                              \\ \cmidrule(l){3-6} 
                      &                            & \multicolumn{1}{c|}{\multirow{3}{*}{EWGS}}                                     & EWGS                                                       & 93.59          & -0.25                                                          \\
                      &                            & \multicolumn{1}{c|}{}                                                          & STE                                                        & 93.52          & -0.32                                                          \\
                      &                            & \multicolumn{1}{c|}{}                                                          & \textbf{Ours}                                                       & \textbf{93.73} & \textbf{-0.11}                                                 \\ \bottomrule
\end{tabularx}
\end{table}

\begin{table}[t]
\scriptsize
\centering
\caption{Top-1 test accuracy (\%) on ImageNet.
``STE (Orig. with FP: 71)'' denotes the accuracy cited from their original paper~\cite{choi2018pact} where the full precision model's accuracy is 71\%.}
\label{tab:results_imagenet}
\setlength{\tabcolsep}{3.pt}
\vspace{-10pt}
\begin{tabular}{@{}c|c|cc|cc@{}}
\toprule
Bit-Width             & Model                      & \multicolumn{1}{c|}{\begin{tabular}[c]{@{}c@{}}Forward \\ Method\end{tabular}} & \begin{tabular}[c]{@{}c@{}}Backward \\ Method\end{tabular} & Accuracy       & \begin{tabular}[c]{@{}c@{}}Accuracy \\ Difference\end{tabular} \\ \midrule
\multirow{5}{*}{W4A4} & \multirow{5}{*}{ResNet-18} & \multicolumn{2}{c|}{Full Precision (FP)}                                                                                                    & 69.80           & -                                                              \\ \cmidrule(l){3-6} 
                      &                            & \multicolumn{1}{c|}{\multirow{4}{*}{PACT}}                                     & STE (Orig. with FP: 71)                           & 69.20           & -1.80                                                           \\
                      &                            & \multicolumn{1}{c|}{}                                                          & STE                                                        & 69.56          & -0.24                                                          \\
                      &                            & \multicolumn{1}{c|}{}                                                          & EWGS                                                       & 69.53          & -0.27                                                          \\
                      &                            & \multicolumn{1}{c|}{}                                                          & \textbf{Ours}                                                       & \textbf{70.01} & \textbf{+0.21}                                                           \\ \bottomrule
\end{tabular}
\end{table}

\subsection{Ablation Study}

\boldhdr{Effect of Training Time}
As shown in Table~\ref{tab:ablation_training_epochs}, both our method and STE benefit from a longer training time, whereas EWGS does not show such improvement. 
Specifically, training 3$\times$ longer increases the top-1 test accuracy of our method and STE by 0.57\% and 0.82\%, respectively, on 2-bit ResNet-20 with CIFAR-10. 
In contrast, extending the training time for EWGS leads to a 0.19\% accuracy drop.

\boldhdr{Effect of Precision Replacement Scheduler}
Table~\ref{tab:ablation_p_scheduler} shows the top-1 test accuracy of 2-bit ResNet-20 trained using constant, linear, exponential, cosine, and logarithmic replacement schedulers as well as a basic baseline without any scheduler, on CIFAR-10.
Figure~\ref{fig:replace_curves_p} illustrates the replacing curves of each scheduler. 
The proposed log-curriculum based replacement scheduler achieves the highest accuracy, outperforming the none-scheduler baseline by 0.55\% and the other four schedulers by 0.36\% to 1.36\%.

\begin{figure}[t]
  \centering
  \begin{subfigure}{.45\linewidth}
    \centering
      \includegraphics[width=4.0cm]{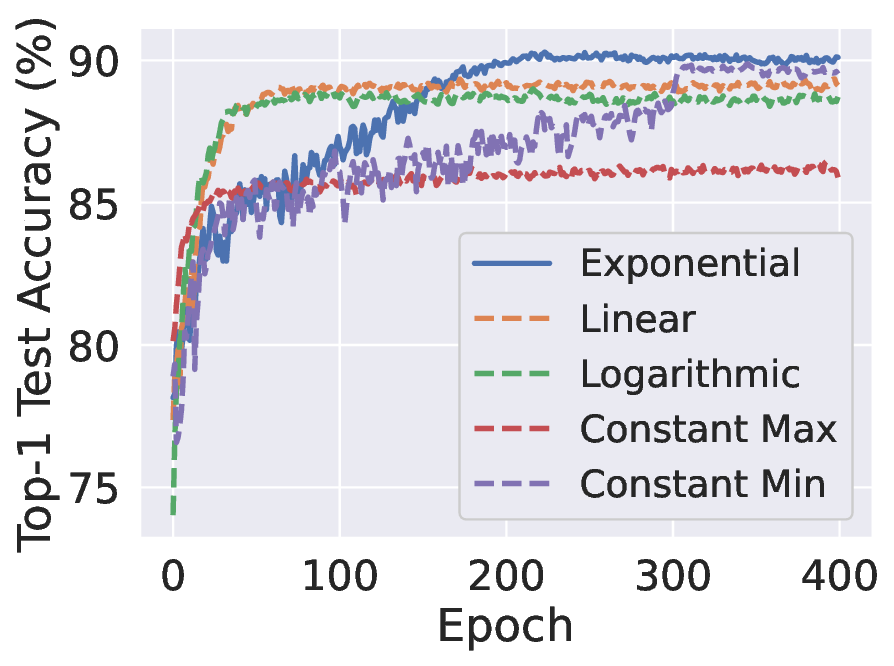}
    \caption{Top-1 test accuracy}
    \label{fig:ablation_u_accuracy}
  \end{subfigure}
  \begin{subfigure}{.45\linewidth}
    \centering
      \includegraphics[width=4.cm]{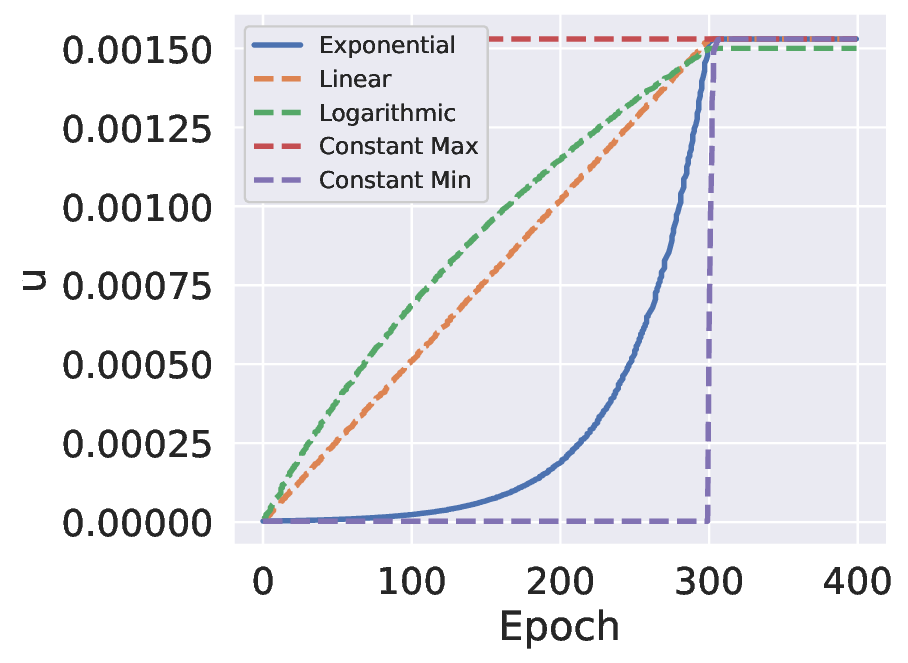}
    \caption{Scheduling curves}
    \label{fig:ablation_u_values}
  \end{subfigure}
  \caption{Illustration of a) top-1 test accuracy evolution of ResNet-20 (W2A2) on CIFAR-10, using five scheduling strategies including constant, linear, logarithmic, and exponential strategies and b) the corresponding scheduling curves for each strategy.}
   \vspace{-10pt}
\end{figure}

\boldhdr{Effect of $\mu$ Scheduler}
Figure~\ref{fig:ablation_u_accuracy} shows the top-1 test accuracy evolution of ResNet-20 quantized to 2-bit weights and activations on CIFAR-10, trained with different $\mu$ scheduling strategies, including constant, linear, logarithmic, and exponential strategies. 
Figure~\ref{fig:ablation_u_values} shows the corresponding value changes for each strategy. 
Exponential $\mu$ Scheduler achieves the best accuracy, aligning with the behavior commonly observed in  Alternating Direction Method of Multipliers (ADMM)-based optimization. 

\begin{table}[t]
\scriptsize
\centering
\caption{Top-1 test accuracy (\%) of different training epochs on CIFAR-10. The forward method is EWGS.}
\label{tab:ablation_training_epochs}
\vspace{-10pt}
\begin{tabularx}{0.48\textwidth}{@{}c|c|cccc@{}}
\toprule
Model                                                                            & \begin{tabular}[c]{@{}c@{}}Training \\ Epochs\end{tabular} & STE          & EWGS         & \textbf{Ours}         \\ \midrule
\multirow{2}{*}{\begin{tabular}[c]{@{}c@{}}ResNet-20\\ (FP: 91.95)\end{tabular}} & 400             & 90.35        & 90.84        & \textbf{91.05}        \\
                                                                                 & 1200            & 91.17 (0.82 {\color{green} \(\uparrow\)}) & 90.65 (0.19 {\color{red} \(\downarrow\)}) & \textbf{91.62 (0.57 {\color{green} \(\uparrow\)})} \\ \bottomrule
\end{tabularx}
\end{table}

\begin{table}[t]
\scriptsize
\centering
\caption{Top-1 test accuracy (\%) of different precision replacement schedulers on CIFAR-10.}
\label{tab:ablation_p_scheduler}
\vspace{-10pt}
\setlength{\tabcolsep}{4.5pt}
\begin{tabularx}{0.48\textwidth}{@{}ccccccc@{}}
\toprule
Model                                                           & None  & Linear & \begin{tabular}[c]{@{}c@{}}Constant \\ (p=0.8)\end{tabular} & Exponential & Cosine & \textbf{Logarithmic} \\ \midrule
\begin{tabular}[c]{@{}c@{}}ResNet-20\\ (FP: 91.95)\end{tabular} & 90.35 & 90.54  & 89.82                                                       & 90.32       & 89.54  & \textbf{90.9}        \\ \bottomrule
\end{tabularx}
\end{table}

\section{Conclusions}
This paper proposes Progressive Element-wise Gradient Estimation (PEGE), a novel and generalizable backward propagation method for Quantization-Aware Training (QAT).
PEGE introduces a logarithmic curriculum-based mixed-precision replacement strategy and an adaptive gradient estimation mechanism, allowing for more effective learning of quantized models.
PEGE is simple to implement, compatible with diverse forward quantizers, and introduces negligible computational overhead, making it highly practical and scalable.
We comprehensively evaluate PEGE across multiple datasets and architectures, showing that it consistently outperforms state-of-the-art backprogation methods in QAT, and even enables low-precision models to match or exceed the performance of their full-precision counterparts.

{
    \small
    \bibliographystyle{ieeenat_fullname}
    \bibliography{main}
}


\end{document}